\documentclass[runningheads]{llncs}
\usepackage[T1]{fontenc}
\usepackage{graphicx}
\usepackage{booktabs}
\usepackage[misc]{ifsym}
\usepackage[strings]{underscore}
\newcommand{\corr}{(\Letter)}
\usepackage{xcolor}
\graphicspath{../Images}
\usepackage{tabularx}
\usepackage{booktabs}
\usepackage{ragged2e}
\usepackage{amsmath}
\begin{document}

\title{From Speech to Subtitles: Evaluating ASR Models in Subtitling Italian Television Programs}
\tocauthor{Alessandro Lucca, Francesco Corso, Francesco Pierri}
\toctitle{From Speech to Subtitles: Evaluating ASR Models in Subtitling Italian Television Programs}
\titlerunning{From Speech to Subtitles}

\author{Alessandro Lucca\inst{1}\and Francesco Corso\inst{1} \and
Francesco Pierri\inst{1} \corr}

\authorrunning{A. Lucca et al.}

\institute{DEIB, Politecnico di Milano, Milano, Italy 
\email{\{alessandro.lucca, francesco.corso, francesco.pierri\}@polimi.it}}

\maketitle              

\begin{abstract}
Subtitles are essential for video accessibility and audience engagement. 
Modern Automatic Speech Recognition (ASR) systems, built upon Encoder–Decoder neural network architectures and trained on massive amounts of data, have progressively reduced transcription errors on standard benchmark datasets. 
However, their performance in real-world production environments, particularly for non-English content like long-form Italian videos, remains largely unexplored.
This paper presents a case study on developing a professional subtitling system for an Italian media company.
To inform our system design, we evaluated four state-of-the-art ASR models (Whisper Large v2, AssemblyAI Universal, Parakeet TDT v3 0.6b, and WhisperX) on a 50-hour dataset of Italian television programs. 
The study highlights their strengths and limitations, benchmarking their performance against the work of professional human subtitlers.
The findings indicate that, while current models cannot meet the media industry's accuracy needs for full autonomy, they can serve as highly effective tools for enhancing human productivity. 
We conclude that a human-in-the-loop (HITL) approach is crucial and present the production-grade, cloud-based infrastructure we designed to support this workflow.
\end{abstract}

\section{Introduction}
\label{sec:introduction}
The production of accurate subtitles has become increasingly important across a wide range of video content, from social media to broadcast television \cite{Gernsbacher2015VideoCaptions}.
High-quality subtitles not only enhance accessibility for individuals with hearing impairments but also support viewer engagement more broadly \cite{WalkerLezicBacic2025}. 
Yet subtitle creation remains a demanding task for human operators, especially in long-form television content where reliability, timing precision, and consistency are critical.

On the technological side, advances in large-scale datasets and End-to-End Neural Network architectures have substantially improved Automatic Speech Recognition (ASR) performance in speech-to-text transcription tasks \cite{prabhavalkar2023endtoendspeechrecognitionsurvey}.
However, ASR research typically evaluates systems using only Word Error Rate (WER), which, while being an important metric for transcription accuracy, it neglects temporal alignment, an essential aspect of subtitle production.
In addition, most benchmarks target English and short audio segments, limiting their applicability to real-world subtitling workflows.

\noindent These gaps are particularly relevant in industrial media contexts, where ASR systems must operate on long-form, noisy, and linguistically diverse broadcast material. To address these challenges, we articulate our contributions through three primary research questions:
\begin{itemize}
    \item[RQ1:] \textbf{How do state-of-the-art ASR models perform across different television genres when evaluated on long-form content?} We addressed this by benchmarking four state-of-the-art models—Whisper Large V2 \cite{radford2022robustspeechrecognitionlargescale}, NVIDIA Parakeet TDT 0.6b V3 \cite{sekoyan2025canary1bv2parakeettdt06bv3efficient}, WhisperX \cite{bain2023whisperxtimeaccuratespeechtranscription}, and AssemblyAI Universal\cite{ramirez2024anatomyindustrialscalemultilingual}—across diverse program formats, including Talk Shows, Investigative Journalism, and Scientific Communication. Our findings indicate that performance is highly genre-dependent; for instance, ``Scientific Communication" achieved consistent, high accuracy with a Word Error Rate (WER) as low as 0.2, while more spontaneous formats like Talk Shows showed significant variability. Our evaluation extends beyond WER to include semantic metrics (BLEURT\cite{sellam2020bleurtlearningrobustmetrics}), subtitle-oriented metrics (Subtitle Edit Rate, SubER \cite{wilken2022subermetricautomaticevaluation}), and entity-level accuracy (Entity Error Rate, EER). We also compute WER on one-minute segments to identify characteristics of video clips that induce higher error rates. 
    \item[RQ2:]\textbf{Can general-purpose LLMs effectively serve as zero-shot reviewers?} We investigated the efficacy of Gemini-2.5-Flash and Gemma3-12b-it as zero-shot reviewers for punctuation restoration and entity correction. The analysis revealed that Gemini-2.5-Flash provided a statistically significant improvement in entity accuracy, reducing the EER by an average of 2.3\%, whereas the smaller Gemma model did not yield significant gains.
    
    \item[RQ3:]\textbf{What are the primary technical and economic trade-offs when deploying an in-house subtitling pipeline versus using proprietary API services?} We addressed the practicalities of deployment by designing a cloud-based infrastructure and conducting a comparative cost-benefit analysis between proprietary APIs and in-house GPU solutions.
    Our findings suggest that while API-based services offer convenience, an internal pipeline utilizing WhisperX on NVIDIA T4/L4 GPUs is more cost-effective for large-scale production, reducing estimated hourly costs from \$7.50 to approximately \$2.56.
\end{itemize}



\section{The ASR systems}
\label{sec:background}

The current Automatic Speech Recognition (ASR) systems are all based on a End To End architecture with two main blocks, an Encoder and a Decoder \cite{prabhavalkar2023endtoendspeechrecognitionsurvey}. The Encoder takes in input the acoustic representation of the audio, typically in log-mel features, and projects it in a very high dimensional space to capture all the information. The Decoder is responsible for the temporal alignment between the acoustic features in the input and the shorter sequence of output labels.
Whisper Large v2 \cite{radford2022robustspeechrecognitionlargescale} is designed with a Transformer Encoder-Decoder architecture, trained as a sequence-to-sequence system on 680k hours of weakly supervised and labelled audio data. WhisperX \cite{bain2023whisperxtimeaccuratespeechtranscription} is a variation of Whisper designed for efficient long-form transcription processing, employing an intelligent preprocessing using Voice Activity Detector (VAD) and Cut\&Merge strategy, and a postprocessing with a phoneme alignment model to ensure maximum temporal alignment. AssemblyAI Universal \cite{ramirez2024anatomyindustrialscalemultilingual} is a 600-million-parameter Conformer Encoder coupled with an RNN-T Decoder, trained on a dataset composed of 12.5M hours of unsupervised data, 188k hours of supervised data, and 1.6M hours of pseudo-labelled data. To handle long-form transcriptions, it uses a divide and conquer strategy, with intelligent segmentation of the input to allow for parallel inference. NVIDIA Parakeet TDT 0.6b v3 \cite{sekoyan2025canary1bv2parakeettdt06bv3efficient} is a FastConformer Encoder paired with a TDT (Token and Duration Transducer) Decoder, a combination specifically designed to achieve a balance between transcription accuracy and computational efficiency. The model has been trained on a 660k-hour subset of the Granary dataset. 

\section{Related Work}
\label{sec:related-work}

Assessing the quality of out-of-the-box ASR transcription is a field of research explored in various dimensions. Kuhn et al. \cite{Kuhn_2023} assessed the performance of multiple open-source and commercial ASR systems using a dataset composed of higher education lectures in English and German, employing the Word Error Rate as the sole evaluation metric. Their findings reveal that the quality promised by vendors is only partially confirmed in practice, with significant variability observed across systems, speakers, and languages.
Similarly, Ferraro et al. \cite{ferraro2023benchmarking} conducted a comparative analysis of a broad range of ASR systems across heterogeneous datasets. Their results demonstrate that ASR performance, again measured through WER, is strongly dependent on the nature and composition of the dataset under examination. In the specific domain of automatic subtitling for professional broadcasting, Davitti et al. \cite{davitti2024using} investigated the readability and quality of subtitles produced by two professional ASR systems, using a semantic metric (NER) for evaluation. Their experiments, conducted on two distinct audiovisual contents, a British talk show and a U.S. film dubbed into Italian, demonstrated that the ASR-generated subtitles consistently fell short of the 98\% accuracy threshold required in the broadcasting industry. In contrast, Romero-Fresco et al. \cite{fit-for-what-purpose-romero-fresco} evaluated the Lexi ASR system using the same NER metric and provided the first certification demonstrating that an ASR system can exceed the 98\% accuracy benchmark on specific types of videos, such as single-speech videos. However, previous studies do not provide a comprehensive evaluation of multiple ASR systems within the specific context of Italian-language subtitling. Most existing work focuses on either a single model, on short and controlled audio samples, or on English-language benchmarks, thus overlooking the challenges posed by real-world audiovisual productions. 
This research aims to bridge these gaps by systematically benchmarking several ASR systems on a medium-scale dataset of real Italian television content. The evaluation covers a wide range of program types and acoustic conditions, providing a more realistic assessment of ASR performance in professional subtitling scenarios.

Several studies have investigated the use of Large Language Models as reviewers within ASR correction pipelines. Ma et al.~\cite{ma2025asrerrorcorrectionusing} show that a zero-shot corrector (GPT-3.5) struggles to enhance transcription accuracy when limited to the 1-best hypothesis, while Fedorchenko et al.~\cite{fedorchenko2025optimizingestoniantvsubtitles} report improvements in the same setting for Estonian language subtitles by using GPT-4o. Li et al.~\cite{Li2024ASRCorrection} demonstrate that a multilingual LLM, when fine-tuned for error correction (Qwen1.5-7b), can substantially enhance ASR output relative to a general-purpose model. Wei et al.~\cite{wei-etal-2025-asr} focus on LLM correction for ASR transcripts in Chinese language: their extensive experiments reveal that prompting is not effective for ASR error correction, finetuning is effective only for a portion of LLMs, while Multi-modal augmentation is the most effective method for error correction, achieving state-of-the-art performance. In any case, these studies do not focus on the Italian language, nor do they consider metrics such as Entity Error Rate to assess the specific improvements that LLMs can bring to entity recognition.

\section{Data and Methods}
\label{sec:data-methods}

\subsection{Dataset}
\label{dataset}
The dataset consists of 50 hours of audio taken from 30 episodes across 5 programs broadcast on Italian public television, each paired with its professional, human-produced SubRip Text subtitle file. 
All episodes were collected from the official website of Radiotelevisione Italiana (RAI), which makes this content publicly available. 
The programs were selected to cover a broad range of television formats and to capture different linguistic and acoustic conditions encountered in real subtitling workflows. 
They fall into three categories: two Talk Shows, two Investigative Journalism programs, and one Scientific Communication program. 
An overview of the dataset characteristics is provided in Tables~\ref{table:reference-dataset} and \ref{table:reference-dataset-ext}.

\begin{table*}[!t]
    \centering    
    \begin{tabular}{|p{3cm}|c|p{2.5cm}|c|c|}
    \hline    
    \textbf{Typology} & \textbf{Number of Episodes} & \textbf{Total Duration (min)} \\
    \hline \hline
    \textbf{Talk Show} & 10 & 664.66 \\ \hline    
    \textbf{Investigative Journalism} & 10 & 1084.66 \\ \hline
    \textbf{Scientific Communication} & 10 & 1,255.53 \\ \hline    
    \end{tabular}
    \\[10pt]
    \caption{Descriptive statistics of the reference dataset considered in the research.}
    \label{table:reference-dataset}
\end{table*}

\subsection{Evaluation metrics}
\label{sec:eval-metric}

The primary evaluation metric used in this study is Word Error Rate (WER), computed over the entire transcript. 
Since the reference subtitles are human-edited rather than verbatim transcriptions, WER should be interpreted with caution: professional captioners often rephrase content to meet readability and timing constraints, introducing a systematic deviation between the audio and the reference text. 
To identify segments where the ASR systems struggle most, WER is also computed on one-minute windows.

In addition to WER, we include SubER \cite{wilken2022subermetricautomaticevaluation}, a syntactic metric specifically designed for subtitling. SubER provides insight into segmentation and timestamp accuracy, both of which are essential for broadcast-quality subtitles.

To assess semantic coherence between reference and hypothesis, we compute AS-BLEURT scores. 
Sentence alignment is performed using the Levenshtein algorithm, ensuring that semantically comparable units are evaluated consistently.

Following the guidelines of major Italian broadcasters~\cite{rai-guidelines}, the study also evaluates three key readability metrics. 
The Number of Characters per Segment (NCS) should fall between 30 and 74; the Mean Segment Duration (MSD) ideally ranges from 1 to 6 seconds; and the Characters per Second (CPS) should remain within the acceptable range of 9 to 15. 
These metrics allow us to assess how well automatically generated subtitles conform to industry-standard readability constraints.

Entity recognition capabilities are evaluated through a dedicated pipeline. 
Named entities are extracted from the reference transcript using the spaCy library and categorized into \textit{Person}, \textit{Organization}, and \textit{Location}.
For each reference entity, we search for the most semantically similar token in the ASR-generated transcript within a temporal window of $\pm$5 seconds, mitigating alignment mismatches between the two transcripts. 
This mapping enables the computation of the Entity Error Rate (EER).

Finally, we investigate the extent to which large language models can improve ASR outputs through post-editing. 
Two LLMs are evaluated in a zero-shot setting for punctuation restoration and entity correction. 
The first is a commercial general-purpose model (Gemini-2.5-Flash), which processes batches of up to 40 subtitles at a time and returns a structured output with the same number of segments. 
The second is a smaller open-source model (Gemma3-12b-it), which does not support structured outputs; therefore, each subtitle is processed individually. 
This analysis provides insight into the potential of LLM-assisted post-processing to enhance subtitle quality.

\subsection{Experimental settings}
\label{subsec:asr-run}
Whisper Large V2, WhisperX, and NVIDIA Parakeet TDT 0.6b V3 are open-source models and are executed in a Google Cloud environment equipped with an NVIDIA T4 GPU. AssemblyAI Universal, being a proprietary system, is accessed via API calls. 
Outputs from all systems are then standardized into a unified JSON structure organized at the segment level. 
For each segment, the JSON file records the transcribed text, start and end timestamps, and the word-level information, including both token text and precise timing.
We plan to publicly release the code used to run and evaluate the models.

\section{Experimental Results}
\label{sec: results}
\begin{figure}[!t]
    \centering
    \includegraphics[width=0.75\textwidth]{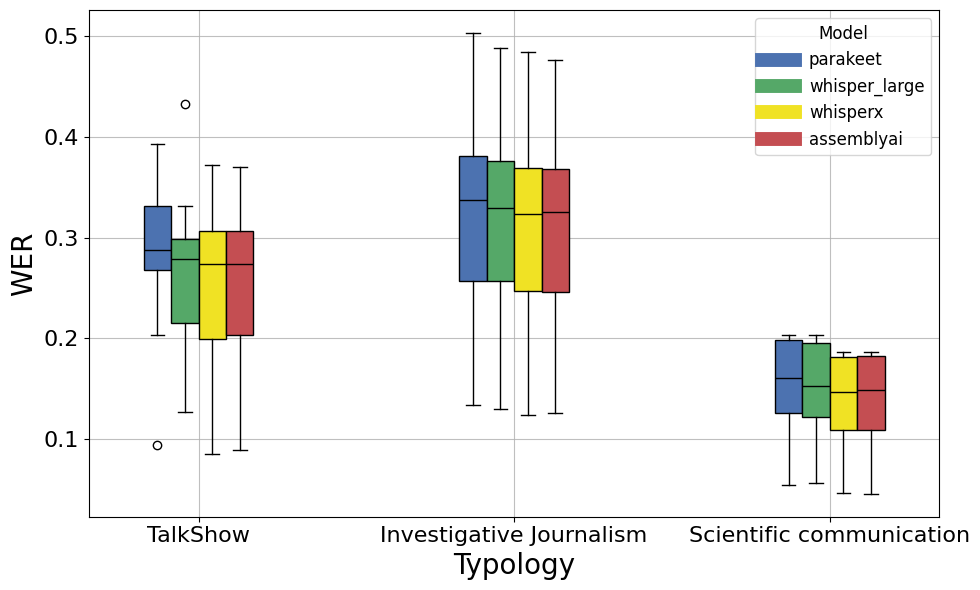}
    \caption{Word Error Rate distribution across different types of TV programs and ASR Models (lower is better).}
    \label{fig:wer}
\end{figure}

The result of the Word Error Rate evaluation is reported in Figure~\ref{fig:wer}. The transcription quality of the "Scientific Communication" program is the most consistent and closely aligned with the ground truth, with a maximum WER of 0.2, a remarkably low error rate, especially considering the potential noise introduced by human post-editing. In contrast, "Investigative Journalism" and "Talk Show" formats exhibit a much wider range of WER values, strongly dependent on the specific episode, thus indicating a higher degree of variability. Across models, all four ASR systems demonstrate comparable overall performance, though WhisperX and AssemblyAI achieve the most accurate results.

\begin{figure}[!t]
    \centering
    \includegraphics[width=0.75\textwidth]{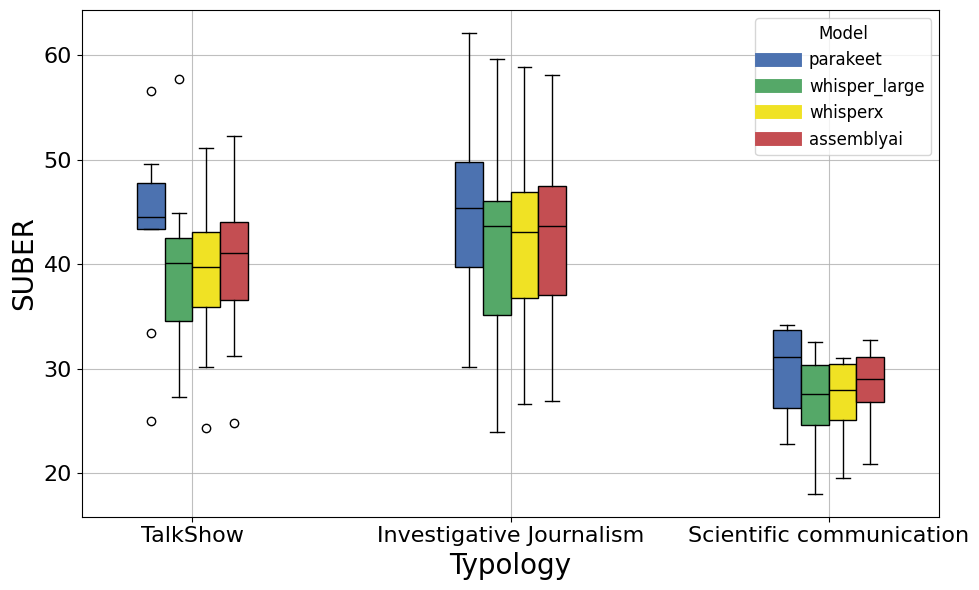}
    \caption{Subtitle Edit Rate distribution across different types of TV programs and ASR Models (lower is better).}
    \label{fig:suber}
\end{figure}

\begin{figure}[!t]
    \centering
    \includegraphics[width=0.75\textwidth]{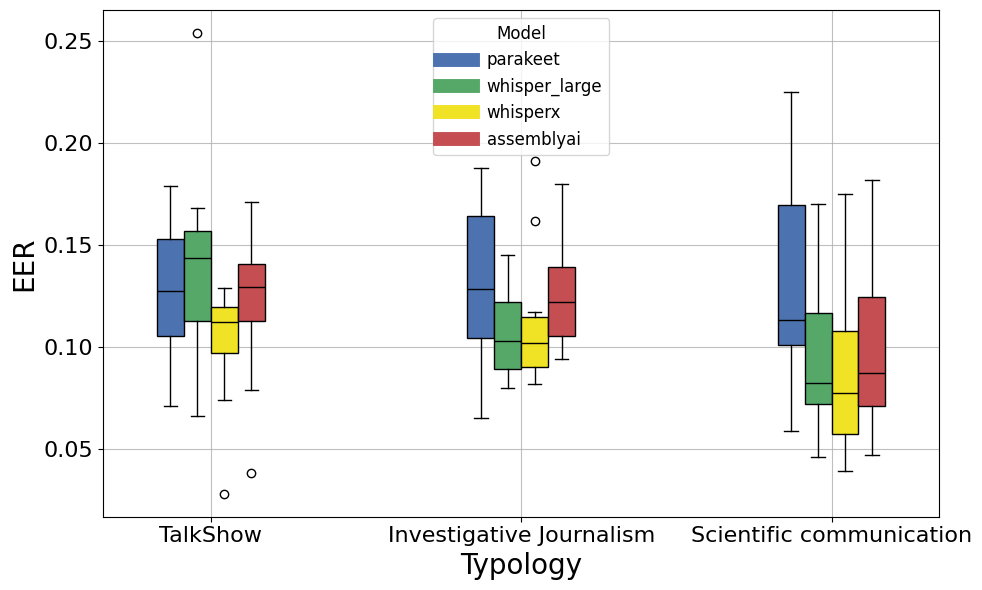}
    \caption{Entity Error Rate distribution across different types of TV programs and ASR Models (lower is better).}
    \label{fig:eer}
\end{figure}

The computation of WER by segmenting the audio in one minute long clips reveals that the ASR struggle in the distinction between speech and singing, trying to transcribe songs, clearly something not desired. The SubER (Figure ~\ref{fig:suber}) and AS-BLEURT results show a strong correlation with WER, indicating that these metrics do not provide additional insights beyond those already captured by WER. However, they confirm the strong dependence of transcription performance on the specific type of program, with the "Scientific Communication" program emerging as the most accurately transcribed program. Concerning the Entity Error Rate (EER), the analysis is conducted on 13,463 extracted entities. Observing Figure~\ref{fig:eer}, a difference emerges among the evaluated models in their ability to recognize entities. Even if all ASR systems achieve a mean EER between 10\% and 15\%, WhisperX consistently achieves the lowest EER in nearly all episodes, establishing itself as the leading model in this task.

\begin{figure}[!t]
    \centering
    \includegraphics[width=0.75\textwidth]{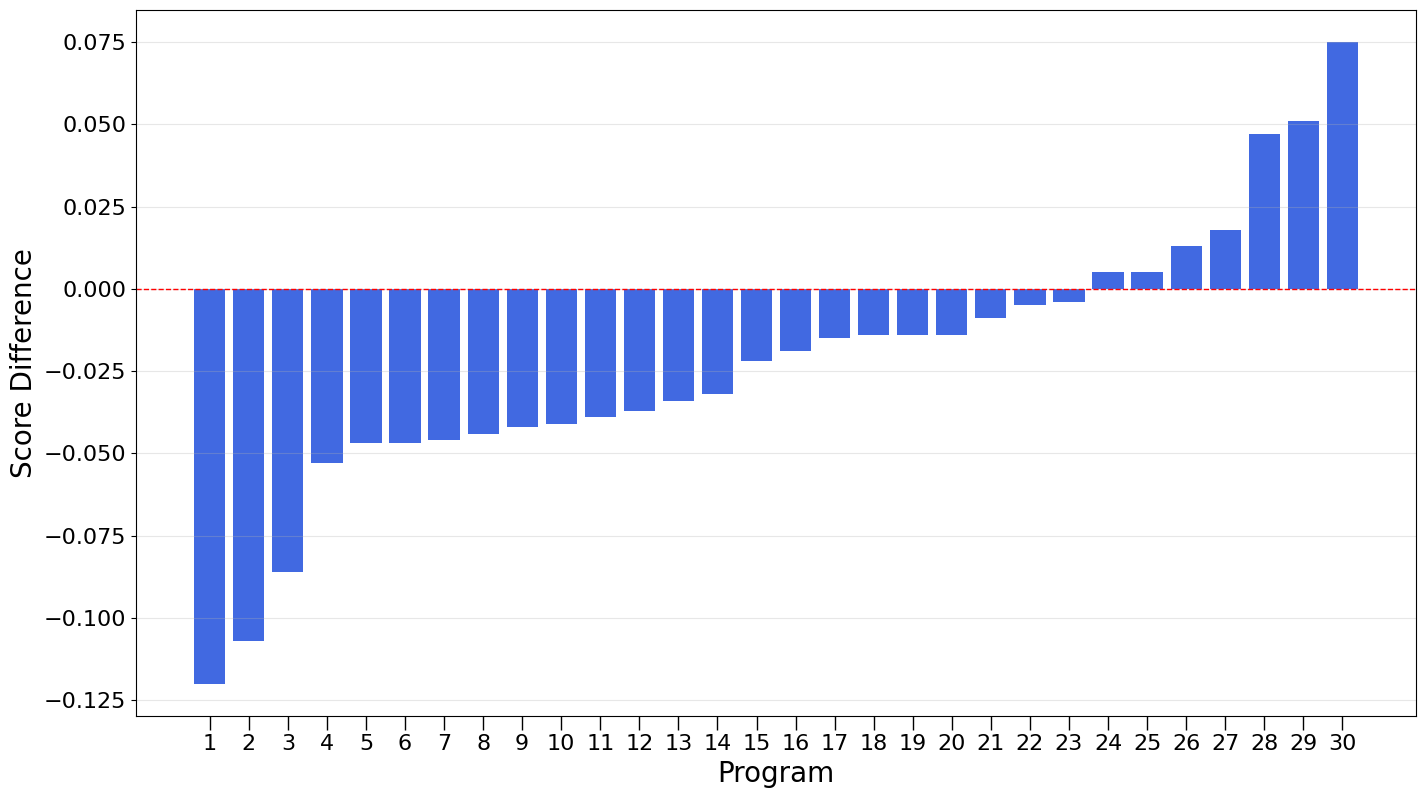}
    \caption{Performance variation after the application of Gemini-2.5-Flash as Reviewer (negative values indicate improvement). Videos are sorted in increasing order by score difference.}
    \label{fig:eer-agent}
\end{figure}

When prompted for entity correction in a zero-shot approach, Gemini-2.5-Flash leads to a statistically significant improvement in EER performance (p-value 0.006), with a mean gain of 2.3\% overall (Figure~\ref{fig:eer-agent}). An example with the differences between the ground-truth, the raw transcription made with WhisperX and the transcription after the revision with Gemini-2.5-Flash can be found in Table~\ref{table:presadiretta-entities}
Using Gemma3-12b-it as a reviewer does not lead to a significant improvement in the entity correction task (p-value 0.542), with a mean gain of 0.4\%.

\definecolor{correct}{RGB}{0,150,0}
\definecolor{wrong}{RGB}{200,0,0}

\begin{table*}[!t]
\centering
\renewcommand{\arraystretch}{1.4}
\begin{tabular}{|p{0.3\textwidth}|p{0.3\textwidth}|p{0.3\textwidth}|}
\hline
\textbf{Reference} & \textbf{WhisperX Transcription} & \textbf{Gemini-2.5-Flash Review} \\
\hline
E' sicuramente una grandissima operazione, sicuramente abbiamo azzerato i vertici della \textbf{'ndrangheta} della provincia di \textbf{Vibo Valentia}, oltre 13 locali in provincia di \textbf{Vibo Valentia} che erano ramificati in tutta \textbf{Italia}. 
&
Sicuramente è una grandissima operazione, sicuramente abbiamo azzerato i vertici dell'\textcolor{wrong}{Andrangheta} della provincia di \textcolor{wrong}{Vibo-Valenzia}. Oltre 12-13 locali di \textcolor{wrong}{Andrangheta} in provincia di \textcolor{wrong}{Vibo-Valenzia} che avevano ramificazioni in tutta \textcolor{correct}{Italia}.
&
Sicuramente è una grandissima operazione, sicuramente abbiamo azzerato i vertici dell'\textcolor{correct}{'ndrangheta} della provincia di \textcolor{correct}{Vibo Valentia}. Oltre 12-13 locali di \textcolor{correct}{ndrangheta} in provincia di \textcolor{correct}{Vibo Valentia} che avevano ramificazioni in tutta \textcolor{correct}{Italia}. \\
\hline
\end{tabular}
\caption{Comparison of Entity Recognition capabilities for a clip. Reference, transcription and review are shown in Italian; colors indicate correct (green) and incorrect (red) entities. The reference is human-edited and may differ from a verbatim transcript.}
\label{table:presadiretta-entities}
\end{table*}

The analysis related to the readability metrics shows that the output of the ASR transcription can not be used "as is", because it fails to respect the necessary compliance guidelines, especially exceeding frequently the NCS metric. However, an intelligent segmentation algorithm that leverages the word timestamp information is sufficient to rebuild a SubRip Text file that is compliant with all the guidelines. Specifically, the segmentation algorithm ensures that each subtitle segment respects the readability constraints. When these are not met, the algorithm automatically splits the text into two lines or multiple segments, prioritizing breaks at punctuation marks. With this strategy, the resulting subtitles comply fully with the main readability metrics considered.

\section{Times and Costs}
The Inverse Real Time Factor (RTFx) has been computed for all four ASR models. Across the 30 programs, Whisper Large V2 achieves a mean Inverse RTFx of 4.873 (\(\pm 0.596\)), WhisperX reaches 14.081 (\(\pm 1.102\)), and AssemblyAI attains 26.035 (\(\pm 4.690\)). For certain samples, Parakeet TDT v3 was executed on a different hardware configuration (NVIDIA L4 GPU instead of T4), resulting in greater variability in its Inverse RTFx values, with a mean of 32.89 and a standard deviation of 16.903; it is included here for completeness. Considering the estimated cost of \$0.64 per hour for using a T4 GPU in the Google Cloud environment, the inference cost for Whisper Large and WhisperX amounts to approximately \$7.04 and \$2.56, respectively. The cost for Parakeet inference varies depending on the L4 GPU usage, while the total cost for AssemblyAI API calls is estimated at \$7.50 (\$0.15 per hour of audio in input). Taken together, the runtime and cost values suggest that in a production environment handling a large volume of videos, it may be preferable to accept longer processing times in exchange for lower operational costs, favoring an internal solution like WhisperX over faster but more expensive external API-based alternatives, thereby optimizing cost-efficiency at scale.
\begin{table*}[!t]
\centering
\caption{Performance Comparison and Cost-Efficiency Analysis of ASR Models. For Inverse RTFx metric, the higher is better.}
\label{table:asr_performance_kpis}
\renewcommand{\arraystretch}{1.4}
\small
\begin{tabularx}
{\textwidth}{X >{\Centering}p{2cm} >{\Centering}p{1.8cm} >{\Centering}p{1.8cm} >{\Centering}p{2.5cm}}
\toprule
\textbf{ASR Model} & \textbf{Estimated Total Inference Cost} & \textbf{Inverse RTFx (Mean)} & \textbf{Std.Dev. ($\pm$)} & \textbf{Hardware Platform} \\
\midrule
Whisper Large V2 & \$7.04   & 4.873  & 0.596  &  NVIDIA T4 \\
WhisperX         & \$2.56   & 14.081 & 1.102  &  NVIDIA T4 \\
AssemblyAI       & \$7.50   & 26.035 & 4.690  & Cloud API \\
Parakeet TDT v3  & Variable & 32.890 & 16.903 & NVIDIA T4/L4 \\
\bottomrule
\addlinespace[0.8ex]
\end{tabularx}
\end{table*}

\section{Discussion}
\label{sec: discussion}
This study demonstrates that integrating an Automatic Speech Recognition system as the central component of a carefully designed subtitling pipeline can significantly streamline the production process. However, achieving fully automated subtitling that meets professional quality standards and industry compliance requirements remains beyond the current technological frontier, as performance varies considerably across program types and contexts. 

Programs with a single speaker, clear articulation, and limited background noise, such as the selected "Scientific Communication" one, can be transcribed with minimal human intervention. In these cases, the application of an appropriate segmentation algorithm ensures compliance with readability and timing guidelines. In contrast, other programs (e.g., the ones of "Investigative Journalism" or "Talk Show") require greater human involvement to preserve the semantic coherence of dialogues and prevent confusion in the transcription, with varying performance between episodes due to differences in linguistic and acoustic conditions. 
Special attention is also required in cases involving songs or foreign languages, where ASR systems tend to diverge from expected behaviour and produce low-quality transcriptions. 

Although all evaluated ASR systems exhibit broadly comparable performance, WhisperX, specifically optimized for long-form transcription, emerges as the most effective model in this benchmarking, particularly for its superior entity recognition capabilities. Moreover, the inclusion of Gemini-2.5-Flash as a reviewer, even in a zero-shot configuration and without "thinking" capabilities, demonstrates a measurable improvement in the correction of misrecognized or misspelled entities. This finding is particularly significant, as in the context of building a production-grade subtitling pipeline for a media company, minimizing entity recognition errors is crucial. Nevertheless, a human-in-the-loop review will remain essential to ensure the highest level of accuracy.

\section{Conclusion and Impact }
\label{sec: conclusion}

This research highlights several key aspects for the industrial development of an infrastructure dedicated to subtitle production in the media industry. 
Our findings suggest that because the various ASR models evaluated exhibit comparable performance, the selection of a system in an industrial setting should move beyond pure accuracy considerations. 
Instead, deployment strategy should be guided by practical trade-offs, such as balancing the operational costs of an in-house GPU-based solution against the convenience of proprietary APIs. 
For instance, handling a large volume of videos favors internal solutions like WhisperX, which optimized cost-efficiency at approximately \$2.56 per hour compared to \$7.50 for external alternatives.

A primary lesson for practitioners is the ``Genre Gap" in ASR reliability.
While programs with clear articulation, such as Scientific Communication, can be transcribed with minimal intervention (WER of 0.2), formats like Talk Shows or Investigative Journalism require significantly greater human involvement to preserve semantic coherence.
Furthermore, we found that raw ASR output is rarely ``broadcast-ready". 
Proper engineering—specifically utilizing word-level timestamps and intelligent segmentation algorithms—is critical to ensure that SubRip files strictly adhere to readability guidelines like the Number of Characters per Segment and Characters per Second.

The integration of a Large Language Model as a zero-shot reviewer represents a significant step forward in automation. 
Specifically, Gemini-2.5-Flash provided a statistically significant improvement (a 2.3\% mean gain) in the correction of misrecognized named entities while preserving the original subtitle structure.
To support these computationally intensive tasks at scale, we developed a decoupled cloud infrastructure that separates the user interface from the processing logic, utilizing NVIDIA L4 GPUs only when required to minimize idle hardware consumption.

While the current system provides a robust foundation, several avenues for future work have been identified to further bridge the gap to full automation. Following recent research in other languages, we plan to investigate multimodal augmentation for the LLM reviewer, providing the model with acoustic features or visual context (e.g., speaker faces or on-screen text) to improve transcription in noisy environments. Moreover, leveraging the existing infrastructure, we aim to extend the pipeline to generate interlingual subtitles and high-quality AI-driven voice dubbing using translation-oriented LLMs and text-to-speech modalities.

Future iterations could incorporate structured evaluations with professional subtitlers to measure the concrete productivity gains and reduced manual effort provided by the HITL pipeline.
In conclusion, hybrid workflows that combine ASR with targeted human review represent the most effective strategy for production environments. 
This human-in-the-loop paradigm could enable scalable automation while maintaining the linguistic and editorial precision required for professional broadcasting.

Consequently, ensuring that end-users have the ability to monitor and edit subtitles in real-time through an intuitive interface remains a crucial component of any successful industrial deployment.

\section{Limitations}
This study presents several limitations that should be considered when interpreting the results.

First, the use of human-made subtitles as reference material introduces an inherent source of evaluation noise. 
These subtitles are not verbatim transcriptions but professionally edited outputs designed for readability and timing. 
As a consequence, the evaluation captures the “distance to industry-compliant output” rather than pure transcription accuracy, and may penalize acceptable paraphrases or differences in segmentation.

Second, the dataset is limited to a small set of television programs from a single Italian broadcaster. 
While this provides a realistic industrial scenario, it restricts generalization to other forms of audiovisual content, such as films or scripted series, which may involve distinct linguistic registers, acoustic environments, and editing styles.

Finally, the study does not incorporate collaborative assessments with professional subtitlers. 
Such evaluations would be crucial for two reasons: (i) they enable a more semantically grounded assessment of ASR errors beyond surface-form mismatches, and (ii) they provide a concrete measure of productivity gains achievable in real production environments. 

\begin{credits}

\subsubsection{\discintname}
The authors have no competing interests to declare that are relevant to the content of this article. 
\end{credits}

%
%
%
%
\bibliographystyle{splncs04}
\bibliography{bibliography.bib} 

@article{davitti2024using,
  title     = {Using ASR Tools to Produce Automatic Subtitles for TV Broadcasting: A Cross-Linguistic Comparative Analysis},
  author    = {Davitti, Elena and Sandrelli, Annalisa and Korybski, Tomasz and Zou, Yining and Orasan, Constantin and Braun, Sabine},
  journal   = {Journal of Audiovisual Translation},
  volume    = {7},
  number    = {2},
  pages     = {1--35},
  year      = {2024},
  doi       = {10.47476/jat.v7i2.2024.305},
  url       = {https://doi.org/10.47476/jat.v7i2.2024.305}
}

@article{fit-for-what-purpose-romero-fresco,
    author = {Romero-Fresco, Pablo and Gauwbergen, Yanou},
    year = {2025},
    month = {01},
    pages = {},
    title = {Fit for What Purpose? NER Certification of Automatic Captions in English and Spanish},
    volume = {15},
    journal = {Applied Sciences},
    doi = {10.3390/app15031387}
}

@article{ferraro2023benchmarking,
  title     = {Benchmarking open source and paid services for speech to text: an analysis of quality and input variety},
  author    = {Ferraro, Antonino and Galli, Antonio and La Gatta, Valerio and Postiglione, Marco},
  journal   = {Frontiers in Big Data},
  volume    = {6},
  pages     = {1210559},
  year      = {2023},
  doi       = {10.3389/fdata.2023.1210559},
  publisher = {Frontiers Media SA},
  url       = {https://doi.org/10.3389/fdata.2023.1210559}
}

@article{Kuhn_2023,
   title={Measuring the Accuracy of Automatic Speech Recognition Solutions},
   volume={16},
   ISSN={1936-7236},
   url={http://dx.doi.org/10.1145/3636513},
   DOI={10.1145/3636513},
   number={4},
   journal={ACM Transactions on Accessible Computing},
   publisher={Association for Computing Machinery (ACM)},
   author={Kuhn, Korbinian and Kersken, Verena and Reuter, Benedikt and Egger, Niklas and Zimmermann, Gottfried},
   year={2023},
   month=dec, pages={1–23} 
}

@misc{wilken2022subermetricautomaticevaluation,
      title={SubER: A Metric for Automatic Evaluation of Subtitle Quality}, 
      author={Patrick Wilken and Panayota Georgakopoulou and Evgeny Matusov},
      year={2022},
      eprint={2205.05805},
      archivePrefix={arXiv},
      primaryClass={cs.CL},
      url={https://arxiv.org/abs/2205.05805}, 
}

@misc{sekoyan2025canary1bv2parakeettdt06bv3efficient,
      title={Canary-1B-v2 \& Parakeet-TDT-0.6B-v3: Efficient and High-Performance Models for Multilingual ASR and AST}, 
      author={Monica Sekoyan and Nithin Rao Koluguri and Nune Tadevosyan and Piotr Zelasko and Travis Bartley and Nikolay Karpov and Jagadeesh Balam and Boris Ginsburg},
      year={2025},
      eprint={2509.14128},
      archivePrefix={arXiv},
      primaryClass={cs.CL},
      url={https://arxiv.org/abs/2509.14128}, 
}

@misc{ramirez2024anatomyindustrialscalemultilingual,
      title={Anatomy of Industrial Scale Multilingual ASR}, 
      author={Francis McCann Ramirez and Luka Chkhetiani and Andrew Ehrenberg and Robert McHardy and Rami Botros and Yash Khare and Andrea Vanzo and Taufiquzzaman Peyash and Gabriel Oexle and Michael Liang and Ilya Sklyar and Enver Fakhan and Ahmed Etefy and Daniel McCrystal and Sam Flamini and Domenic Donato and Takuya Yoshioka},
      year={2024},
      eprint={2404.09841},
      archivePrefix={arXiv},
      primaryClass={eess.AS},
      url={https://arxiv.org/abs/2404.09841}, 
}

@misc{bain2023whisperxtimeaccuratespeechtranscription,
      title={WhisperX: Time-Accurate Speech Transcription of Long-Form Audio}, 
      author={Max Bain and Jaesung Huh and Tengda Han and Andrew Zisserman},
      year={2023},
      eprint={2303.00747},
      archivePrefix={arXiv},
      primaryClass={cs.SD},
      url={https://arxiv.org/abs/2303.00747}, 
}

@inproceedings{radford2022robustspeechrecognitionlargescale,
  title={Robust speech recognition via large-scale weak supervision},
  author={Radford, Alec and Kim, Jong Wook and Xu, Tao and Brockman, Greg and McLeavey, Christine and Sutskever, Ilya},
  booktitle={International conference on machine learning},
  pages={28492--28518},
  year={2023},
  organization={PMLR}
}

@misc{prabhavalkar2023endtoendspeechrecognitionsurvey,
      title={End-to-End Speech Recognition: A Survey}, 
      author={Rohit Prabhavalkar and Takaaki Hori and Tara N. Sainath and Ralf Schlüter and Shinji Watanabe},
      year={2023},
      eprint={2303.03329},
      archivePrefix={arXiv},
      primaryClass={eess.AS},
      url={https://arxiv.org/abs/2303.03329}, 
}

@misc{sellam2020bleurtlearningrobustmetrics,
      title={BLEURT: Learning Robust Metrics for Text Generation}, 
      author={Thibault Sellam and Dipanjan Das and Ankur P. Parikh},
      year={2020},
      eprint={2004.04696},
      archivePrefix={arXiv},
      primaryClass={cs.CL},
      url={https://arxiv.org/abs/2004.04696}, 
}

@misc{ma2025asrerrorcorrectionusing,
      title={ASR Error Correction using Large Language Models}, 
      author={Rao Ma and Mengjie Qian and Mark Gales and Kate Knill},
      year={2025},
      eprint={2409.09554},
      archivePrefix={arXiv},
      primaryClass={cs.CL},
      url={https://arxiv.org/abs/2409.09554}, 
}

@misc{Li2024ASRCorrection,
  author       = {Sheng Li and Chen Chen and Chin Yuen Kwok and Chenhui Chu and Eng Siong Chng and Hisashi Kawai},
  title        = {Investigating ASR Error Correction with Large Language Model and Multilingual 1-best Hypotheses},
  booktitle    = {Proceedings of Interspeech 2024},
  year         = {2024},
  month        = sep,
  pages        = {},
  address      = {Kos, Greece},
  publisher    = {ISCA},
  url          = {https://www.isca-archive.org/interspeech_2024/li24h_interspeech.pdf}
}

@misc{fedorchenko2025optimizingestoniantvsubtitles,
      title={Optimizing Estonian TV Subtitles with Semi-supervised Learning and LLMs}, 
      author={Artem Fedorchenko and Tanel Alumäe},
      year={2025},
      eprint={2501.05234},
      archivePrefix={arXiv},
      primaryClass={cs.CL},
      url={https://arxiv.org/abs/2501.05234}, 
}

@inproceedings{wei-etal-2025-asr,
    title = "{ASR}-{EC} Benchmark: Evaluating Large Language Models on {C}hinese {ASR} Error Correction",
    author = "Wei, Victor Junqiu  and
      Wang, Weicheng  and
      Jiang, Di  and
      Song, Yuanfeng  and
      Wang, Lu",
    editor = "Potdar, Saloni  and
      Rojas-Barahona, Lina  and
      Montella, Sebastien",
    booktitle = "Proceedings of the 2025 Conference on Empirical Methods in Natural Language Processing: Industry Track",
    month = nov,
    year = "2025",
    address = "Suzhou (China)",
    publisher = "Association for Computational Linguistics",
    url = "https://aclanthology.org/2025.emnlp-industry.110/",
    doi = "10.18653/v1/2025.emnlp-industry.110",
    pages = "1567--1575",
    ISBN = "979-8-89176-333-3"
}

@misc{rai-guidelines,
  author       = {{RAI - Radiotelevisione Italiana}},
  title        = {Norme e convenzioni editoriali essenziali sottotitoli televisivi per spettatori sordi e con difficoltà uditive a cura di RAI},
  year         = {2021},
  howpublished = {\url{https://www.rai.it/dl/doc/2020/10/19/1603121663902_PREREGISTR_22_feb_2016.pdf}}
}

@article{Gernsbacher2015VideoCaptions,
  author    = {Gernsbacher, Morton Ann},
  title     = {Video Captions Benefit Everyone},
  journal   = {Policy Insights from the Behavioral and Brain Sciences},
  volume    = {2},
  number    = {1},
  pages     = {195--202},
  year      = {2015},
  month     = oct,
  doi       = {10.1177/2372732215602130},
  pmid      = {28066803},
  pmcid     = {PMC5214590},
  publisher = {SAGE Publications}
}

@inproceedings{WalkerLezicBacic2025,
  author    = {Kaitlin Walker and Ayla Lezic and Dinko Bačić},
  title     = {Unveiling the Impact of Subtitles: Insights into Recall and Viewer Visual Engagement on Streaming Video Content},
  booktitle = {Proceedings of MIPRO 2025 — Human–Computer Interaction (HCI)},
  year      = {2025},
  pages     = {1540--1545},
  address   = {Ohrid, North Macedonia},
  publisher = {MIPRO},
  note      = {Quinlan School of Business, Loyola University Chicago}
}




\clearpage
\section{Appendix}

\begin{figure}[h]
    \centering
    \includegraphics[width=0.9\linewidth]{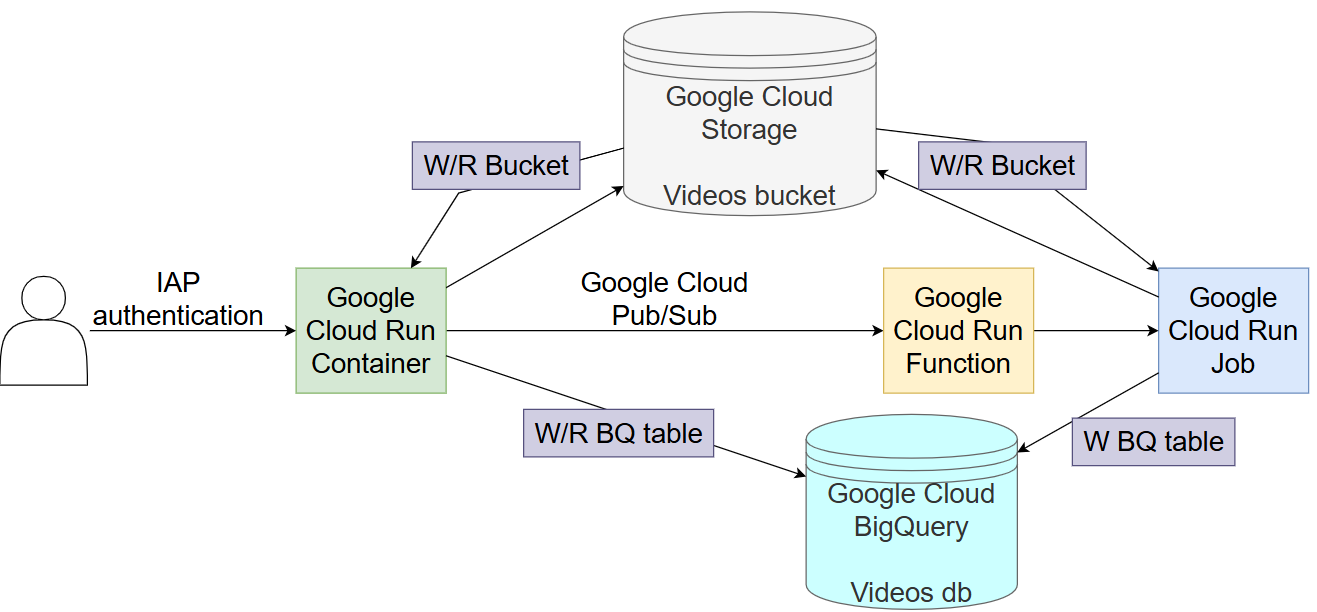}
    \caption{High-level overview of the developed infrastructure.}
    \label{fig:schema}
\end{figure}

The high level schema of the infrastructure currently in production and developed following the results of the research is reported in Figure~\ref{fig:schema}. The subtitle processing pipeline runs on a Google Cloud Run Job, while the user interface is hosted on a Google Cloud Run Service Container. The authentication is handled via Identity-Aware Proxy identification, mantaining an allow-list of authorized users. When a user initiates processing through the UI, the Job is triggered via a dedicated Cloud Run Service Function that listens to messages on a specific Pub/Sub topic. The separation of responsibilities between the components offers two key advantages: first, it enables broader accessibility to the subtitling service, allowing it to be invoked from external entry points without relying solely on the front-end interface; second, it ensures continuous service availability for users while minimizing hardware resource consumption. This is achieved by delegating computationally intensive tasks to the Job instance, which is equipped with an NVIDIA L4 GPU to ensure optimal performance and is automatically activated only when required, shutting down once the processing is complete. Google Cloud Storage and BigQuery are used to manage the data and metadata generated during the workflow.

\begin{table}[h]    
    \centering
    \begin{tabular}{|p{40em}|}
    \hline
    \textbf{System Prompt} \\
    \hline    
    You are a professional subtitle proofreader, specialized in Italian multimedia content. 
    You will receive a series of subtitles from a video segment produced by Italian Media company (e.g. journalism, interviews, sports, etc.). 
    Each subtitle includes an "index" field and a "subtitle" field (in Italian), extracted from an SubRip Text file.
    These subtitles were automatically generated with Whisper and may contain typos, formatting issues, or transcription errors.
    Your task is to correct evident mistakes such as typos, mis-transcribed proper names, or minor grammatical errors.
    Do not alter the structure or content that is already correct.
    IMPORTANT: return only a JSON array with the same number of elements as the input. Each element must be an object with "index" and "correct\_sub". \\\hline     
    \textbf{Requested Structured Output}  \\\hline    
    class CorrectSubs(BaseModel):\\
    \ \ \ \ index: int = Field(..., description="The index of the subtitled sentence")\\
     \ \ \ \ correct\_sub: str = Field(..., description="The corrected sub")\\\hline
    \end{tabular}
    \\[10pt]
    \caption{System prompt and requested structured output for the reviewer LLM.}
    \label{table:reviewer_definition}
\end{table}

\begin{table*}[t]
    \centering    
    \begin{tabular}{|p{3cm}|c|p{2.5cm}|c|c|}
    \hline    
    \textbf{Program} & \textbf{Number of Episodes} & \textbf{Typology} & \textbf{Total Duration (min)}  \\
    \hline \hline
    \textbf{Mezz'ora in più} & 5 & Talk Show & 149.93 \\ \hline
    \textbf{Porta a Porta} & 5 & Talk Show & 514.73 \\ \hline
    \textbf{Presa Diretta} & 5 & Investigative Journalism & 540.86 \\ \hline
    \textbf{Report} & 5 & Investigative Journalism & 543.80 \\ \hline
    \textbf{Ulisse} & 10 & Scientific Communication & 1,255.53 \\
    \hline
    \end{tabular}
    \\[10pt]
    \caption{Descriptive statistics of the reference dataset considered in the research.}
    \label{table:reference-dataset}
\end{table*}


\begin{table}[!t]     
    \centering
    \begin{tabular}{|p{7em} c c c c|}
    \hline
    \textbf{Program} & \textbf{Date} & \textbf{Typology} & \textbf{Duration (min)} & \textbf{Reference Number} \\
    \hline \hline    
    \text{Mezz'ora in più} & 14/03/2021 & Talk Show & 30.2 & 462 \\
    \text{Mezz'ora in più} & 23/05/2021 & Talk Show & 28.83 & 442 \\
    \text{Mezz'ora in più} & 10/10/2021 & Talk Show & 35.25 & 540 \\
    \text{Mezz'ora in più} & 24/10/2021 & Talk Show & 27.65 & 435 \\
    \text{Mezz'ora in più} & 27/03/2022 & Talk Show & 28.00 & 460 \\
    \hline
    \text{Porta a Porta} & 09/06/2021 & Talk Show & 86.93 & 1169 \\
    \text{Porta a Porta} & 15/06/2022 & Talk Show & 104.21 & 1561 \\
    \text{Porta a Porta} & 22/06/2022 & Talk Show & 118.64 & 1824 \\
    \text{Porta a Porta} & 23/06/2022 & Talk Show & 96.25 & 1497 \\
    \text{Porta a Porta} & 28/06/2022 & Talk Show & 108.70 & 1639 \\
    \hline
    \text{Presa Diretta} & 15/03/2021 & Investigative Journalism & 106.29 & 1394 \\
    \text{Presa Diretta} & 07/02/2022 & Investigative Journalism & 105.76 & 1505 \\
    \text{Presa Diretta} & 14/02/2022 & Investigative Journalism & 109.61 & 1766 \\
    \text{Presa Diretta} & 14/03/2022 & Investigative Journalism & 109.67 & 1773 \\
    \text{Presa Diretta} & 13/03/2023 & Investigative Journalism & 109.53 & 1610 \\    
    \hline
    \text{Report} & 12/04/2021 & Investigative Journalism & 110.93 & 1471 \\
    \text{Report} & 10/05/2021 & Investigative Journalism & 106.97 & 1363 \\
    \text{Report} & 20/12/2021 & Investigative Journalism & 112.27 & 1635 \\
    \text{Report} & 11/04/2022 & Investigative Journalism & 107.64 & 1659 \\    
    \text{Report} & 13/06/2022 & Investigative Journalism & 105.99 & 1571 \\   
    \hline
    \text{Ulisse} & 21/04/2021 & Scientific Communication & 126.74 & 1702 \\
    \text{Ulisse} & 28/04/2021 & Scientific Communication & 127.22 & 1670 \\
    \text{Ulisse} & 05/05/2021 & Scientific Communication & 125.93 & 1709 \\
    \text{Ulisse} & 27/05/2021 & Scientific Communication & 122.00 & 1482 \\
    \text{Ulisse} & 03/06/2021 & Scientific Communication & 124.15 & 1514 \\  
    \text{Ulisse} & 25/01/2022 & Scientific Communication & 128.90 & 1586 \\
    \text{Ulisse} & 09/04/2022 & Scientific Communication & 123.96 & 1664 \\
    \text{Ulisse} & 16/04/2022 & Scientific Communication & 127.13 & 1666 \\    
    \text{Ulisse} & 23/04/2022 & Scientific Communication & 127.13 & 1853 \\    
    \text{Ulisse} & 30/04/2022 & Scientific Communication & 122.37 & 1555 \\
    \hline
    \end{tabular}
    \\[10pt]
    \caption{Descriptive statistics of the reference dataset considered in the research.}
    \label{table:reference-dataset-ext}
\end{table}

\end{document}